# Dimension Reduction by Mutual Information Feature Extraction

Ali Shadvar[1]

[1]Department of Biomedical Engineering, Science and Research Branch, Islamic Azad University, Langrud, Iran

`Shadvar@iaul.ac.ir`

## ABSTRACT

*During the past decades, to study high-dimensional data in a large variety of problems, researchers have proposed many Feature Extraction algorithms. One of the most effective approaches for optimal feature extraction is based on mutual information (MI). However it is not always easy to get an accurate estimation for high dimensional MI. In terms of MI, the optimal feature extraction is creating a feature set from the data which jointly have the largest dependency on the target class and minimum redundancy. In this paper, a component-by-component gradient ascent method is proposed for feature extraction which is based on one-dimensional MI estimates. We will refer to this algorithm as Mutual Information Feature Extraction (MIFX). The performance of this proposed method is evaluated using UCI databases. The results indicate that MIFX provides a robust performance over different data sets which are almost always the best or comparable to the best ones*
.

## 1. INTRODUCTION

Dimensionality reduction of the raw input variable space is an essential pre-processing step in the classification process. There are two main reasons to keep the dimensionality of the input features as small as possible: computational cost and classification accuracy. It has been observed that added irrelevant features may actually degrade the performance of classifiers if the number of training samples is small relative to the number of features [1-2].

Reduction of the number of input variables can be done by selecting relevant features (i.e., feature selection) [3-4] or extracting new features containing maximal information about the class label from the original ones (i.e., feature extraction) [5-6]. To keep some of original features it may be more suitable to obtain feature selection, besides when the orders of magnitude makes the irrelevant features outnumber the relevant features it takes a great deal of training data to gain reliable transformation, on the other hand since to switch from a feature to another one is a discrete occurrence then feature selection is not a smooth process. Another reason which motivates using feature extraction over selection is its extraction power on distributing relevant information within different original features which gives us the ability of more compaction [7].

The function which describes the feature extraction is $\mathbf{z} = f(\mathbf{x}), \mathbf{z} \in \mathrm{R}^N$ that can be either linear or nonlinear. This is the use of classifier which determines whether linear or nonlinear extraction method ought to be used or not. For this reason applying whether a nonlinear feature extraction method before a linear classifier or a linear classifier followed by nonlinear one is usually common. In the first case the data is projected by a nonlinear feature extractor on a set of variables in which the nonlinear patterns are unfolded and the separation of the classes will be possible by a linear classifier. In the second case it's the classifier which holds the responsibility of finding the nonlinear separation boundaries [8]. In [9] it has been shown that a proper linear transformation on data specially improves the performance of a simple k-nearest-neighbors (KNN) classifier that's why we will consider the linear feature extraction in this paper.

So far many researchers have presented wide variety of methods for linear feature extraction. That PCA is one of the most famous ones. Finding an orthogonal set of projection vectors for extracting the features is the ultimate goal of PCA which is done by maximizing the variance of data. Though it is a good method to reduce dimensionality but because of its unsupervised nature it is not suitable for feature extraction in classification [10].

Linear discrimination analysis (LDA) is also a well-known and popular linear dimensionally reduction algorithm for supervised feature extraction [11]. LDA computes a linear transformation by maximizing the ratio of between-class distance to within-class distance, thereby achieving maximal discrimination. In LDA, a transformation matrix from an n-dimensional feature space to a d-dimensional space is determined such that the Fisher criterion of between-class scatter over within-class scatter is maximized [12].

However, traditional LDA method is based on the restrictive assumption that the data are homoscedastic, *i.e,* data in which classes have equal covariance matrices. In particular, it is assumed that the probability density functions of all classes are Gaussian with identical covariance matrix but with different means [13]. Moreover, traditional LDA cannot solve the problem posed by nonlinearly separable classes. Hence, its performance is unsatisfactory for many classification problems in which nonlinear decision boundaries are necessary. To solve this, nonlinear extension of LDA has been proposed [14-15].

Moreover, LDA-based algorithms generally suffer from Small Sample Size (SSS) problem when the number of training samples is less than the dimension of feature vectors [16]. A traditional solution to this problem is to apply PCA in conjunction with LDA [17]. Recently, more effective solutions have been proposed to solve the SSS [10].

Another problem that is common to most DA methods is that these methods can only extract $C$-1 features from the original feature space where $C$ is the number of classes [18]. Recently, a method based on DA was proposed, known as Subclass Discriminant Analysis (SDA), for describing a large number of data distributions and to solve the limitation posed by the DA methods in the number of features that can be extracted [19].

One of the most effective approaches for optimal feature extraction is based on (MI). MI measures the mutual dependence of two or more variables. In this context, the feature extraction process is creating a feature set from the data which jointly have largest dependency on the target class and minimal redundancy among themselves. However, it is almost impossible to get an accurate estimation for high-dimensional MI. In [7, 20], a method was proposed, known as MRMI, for learning linear discriminative feature transform using an approximation of the MI between transformed features and class labels as a criterion. The approximation is inspired by the quadratic Renyi entropy which provides a nonparametric estimate of the MI. However, there is no general guarantee that maximizing the approximation of MI using Renyi's definition is equivalent to maximizing MI defined by Shannon. Moreover, MRMI algorithm is subjected to the curse of dimensionality. In [8] a method of extracting of features based on one dimensional MI has been presented which is called MMI, in this method the first feature is extracted in a way that maximizes the MI between the extracted feature and the class of data, the other features must be extracted in a way that first be orthogonal and second maximize the MI between the extracted features and the class label but in general the orthogonality of newly extracted feature over the previous ones cannot guarantee that the new one is independent from the previous ones therefore this method still cannot eliminate redundancy. To overcome the difficulties of MI estimation for feature extraction, Parzen window modeling was also employed to estimate the probability density function [21]. However, Parzen model may suffer from the "curse of dimensionality," which refers to the over fitting of the training data when their dimension is high [8].

The purpose of this paper is to introduce an efficient method to extract features with maximal dependency to the target class and minimal redundancy among themselves using one-dimensional MI estimation to overcome the above mentioned practical obstacle. The proposed method is then evaluated by using six databases. The obtained results by using proposed method is compared with those obtained by using PCA, LDA, SDA [19] and MI-based feature extraction method (MRMI-SIG) proposed in [20]. The results indicate that MIDA provides a robust performance over different data sets with different characteristics which are almost always the best or comparable to the best ones.

The rest of the paper is divided as follows. In Section II, a summary of information theory concepts is provided. In Section III, we describe our algorithm for feature extraction. In Section IV, based on experiments we compare the practical result of our method with other methods. In Section V, we conclude.

## 2. BACKGROUND ON MUTUAL INFORMATION AND FEATURE EXTRACTION

Mutual information is a nonparametric measure of relevance between two variables. Shannon's information theory provides a suitable formalism for quantifying these concepts [22]. Assume a random variable $X$ representing continuous valued random feature vector, and a discrete-valued random variable $C$ representing the class labels. In accordance with Shannon's information theory, the uncertainty of the class label $C$ can be measured by entropy $H(C)$ as

$$H(C) = -\sum_{c \in C} p(c) \log p(c) \tag{1}$$

Where $p(c)$ represents the probability of the discrete random variable $C$. The uncertainty about $C$ given a feature vector $X$ is measured by the conditional entropy as

$$H(C|X) = -\int_x p(x) \left( \sum_{c \in C} p(c|x) \log p(c|x) \right) dx \tag{2}$$

Where $p(c|x)$ is the conditional probability for the variable $C$ given $X$.

In general, the conditional entropy is less than or equal to the initial entropy. The conditional entropy is equal if and only if variables $C$ and $X$ are independent. By definition, the amount that the class uncertainty is decreased by is the MI. As such, $I(X;C) = H(C) - H(C|X)$. After applying the identities $p(c,x) = p(c|x) p(x)$ and $p(c) = \int_x p(c,x) dx$, $I$ can be expressed as

$$I(X;C) = \sum_{c \in C} \int_x p(c,x) \log \frac{p(c,x)}{p(c) p(x)} dx \tag{3}$$

If the MI between two random variables is large, it means two variables are closely related. Indeed, MI is zero if and only if the two random variables are strictly independent [23].

In classification problem suitable features are those which have a higher quantity of MI with regard to the classes. There are two bounds on Bayes error which justify the use of MI for feature extractions. The first one is Hellman and Raviv's upper bound $p_e \leq (H(C) - I(X;C))/2$. The second one is Fano's lower bound $p_e \geq (H(C) - I(X;C) - 1)/\log(N_c)$. As the MI grows the bounds decrease which it is resulted in decreasing of Bayes error. This shows that using MI is a reasonable criterion for feature extraction. On the other hand according to inequality of data processing for any deterministic transformation $T(\cdot)$ we hold

$$I(T(x);C) \leq I(X;C) \tag{4}$$

This equality is only held when the transformation process is invertible [24] so no improvement will occur in MI existing between data and classes, for this reason our objective in this paper is to propose a heuristic method for feature extraction which is based on minimal-redundancy-maximal-relevance framework which maximizes the information in a reduced space.

## 3. MUTUAL INFORMATION FEATURE EXTRACTION

To obtain an optimal extraction of features we need to create a new set of features from the original ones with the largest dependency on the target class. Let us denote by $X$ the original feature set as the sample of continuous-valued random vector, and by discrete-valued random variable $C$ the class labels. The problem is to find a linear mapping $W$ such that the transformed features

$$Y = W^T X \tag{5}$$

Maximize the MI between the transformed features $Y$ and the class labels $C$, $I(W^T X;C)$. That is, we seek

$$W_{opt} = \arg\max_W I(W^T X;C) \tag{6}$$

$$I(Y;C) = \sum_{c \in C} \int \ldots \int p(y_1 \ldots y_m) \log \frac{p(y_1 \ldots y_m, c)}{p(y_1 \ldots y_m) p(c)} \times dy_1 \ldots dy \tag{7}$$

The requirement of the knowledge on underlying probability density functions (pdfs) of the data and the integration on these pdfs always makes it difficult to get an accurate estimation for high-dimensional MI [25]. The above mentioned solution is not practically applicable due to its requirement to enormous computations when it comes to complex problems.

To overcome the above mentioned practical obstacle, we propose a heuristic component-by-component gradient–ascent method for feature extraction that uses a one dimensional MI estimation using histogram method [26] that is a popular way to estimate MI for low-dimensional data space. Histogram estimators can deliver satisfactory results under low-dimensional data spaces. The first feature must be extracted in way that maximizes the MI between the extracted feature and the class label, to achieve this according to figure 1 we should maximize area number two.

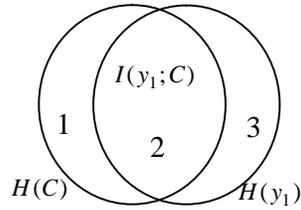

Figure 1. The relation between the feature that will be extracted and output classes

$$w_1 = \arg\max_{w_1} I(w_1^T X; C) \qquad (8)$$

$$y_1 = w_1^T X \qquad (9)$$

$$W_s \leftarrow \{w_1\} \qquad (10)$$

the rest of the features must be extracted the way that at the same time have the maximum relevance with the class label and minimum relevance with the pervious extracted features, to achieve this goal for extracting ith feature in its iteration according to figure 2 area four must be maximized such that $I(w_i^T X | W_s^T X; C)$.

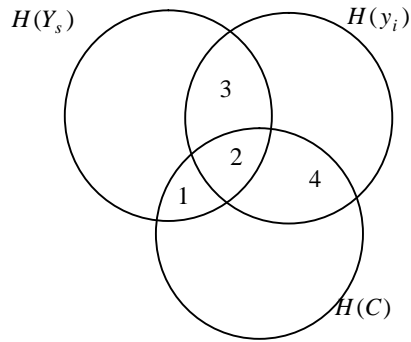

Figure 2. The relation between the feature that will be extracted, previously extracted features and output classes

$$w_i = \arg\max_{w_2} I(w_i^T X | W_s^T X; C) \qquad (11)$$

$$y_i = w_i^T X \qquad (12)$$

$$W_s \leftarrow \{w_i\} \qquad (13)$$

Calculating $I(w_i^T X | W_s^T X; C)$ needs the estimation of high-dimensional MI which is practically impossible. With reformulating this we have

International Journal of Computer Science & Information Technology (IJCSIT) Vol 4, No 3, June 2012

$$w_2 = \arg\max_{w_2} \left\langle I(w_i^T X; C) - I(w_i^T X; W_s^T X; C) \right\rangle \quad (14)$$

$$y_i = w_i^T X \quad (15)$$

In [27] an approximation of $I(w_i^T X; W_s^T X; C)$ based on one dimensional MI has been given with the assumption that the conditioning by the class C does not change. In this paper we have applied this approximation to calculate $I(w_i^T X; W_s^T X; C)$.

$$I(w_i^T X; W_s^T X; C) \approx \sum_{w_s \in W_s} \frac{I(w_i^T X; W_s^T X) I(W_s^T X; C)}{H(W_s^T X)} \quad (16)$$

The rewritten form of this formula goes as follows:

$$I(w_i^T X | W_s^T X; C) \approx \left\langle I(w_i^T X; C) - \sum_{w_s \in W_s} \frac{I(w_i^T X; W_s^T X) I(W_s^T X; C)}{H(W_s^T X)} \right\rangle \quad (17)$$

The first part of (17) indicates the relevancy of the newly extracted feature with the class label And the second part indicates the dependency of the newly extracted feature with the former extracted features, therefore to extract features that having maximum dependency to the class label and minimum relevancy with the pervious extracted features we must extract this new features in a way that minimizes $\sum_{w_s \in W_s} \frac{I(w_i^T X; W_s^T X) I(W_s^T X; C)}{H(W_s^T X)}$ on the condition that maximizes $I(w_i^T X; C) - \sum_{w_s \in W_s} \frac{I(w_i^T X; W_s^T X) I(W_s^T X; C)}{H(W_s^T X)}$. Then to extract the ith feature which occurs in the ith iteration we have

$$w_i = \arg \begin{cases} \min\left\langle \sum_{w_s \in W_s} \frac{I(w_i^T X; W_s^T X) I(W_s^T X; C)}{H(W_s^T X)} \right\rangle \\ \max\left\langle I(w_i^T X; C) - \sum_{w_s \in W_s} \frac{I(w_i^T X; W_s^T X) I(W_s^T X; C)}{H(W_s^T X)} \right\rangle_{w_i} \end{cases} \quad (18)$$

$$W_s \leftarrow \{w_i\} \quad (19)$$

$$y_i = w_i^T X \quad (20)$$

To optimize and learn the linear mapping *W* we apply a genetic algorithm which unlike the similar techniques does not rely on computing local first- or second-order derivatives for guiding the search process; therefore for searching wide solution pace and avoiding local minima GA is generally more flexible than the others. To implement the GA, we use genetic algorithm and direct search toolbox for use in Matlab.

The proposed MI-based feature extraction can be summarized by the following procedure:

International Journal of Computer Science & Information Technology (IJCSIT) Vol 4, No 3, June 2012

1) Initialization:

   Set $X$ to the initial feature set;

   Set $S$ to the empty set;

   Set $W_s$ to the empty set;

   Set t to the desired number of feature that will extract

2) Find the first weighting vector

   $$w_1 = \arg\max_{w_1} I(w_1^T X; C)$$

   $$W_s \leftarrow \{w_1\}$$

3) Finding the other weighting vector s

   For i=1 to t do

   $$w_i = \arg\left\{\min\left\langle \sum_{w_s \in W_s} \frac{I(w_i^T X; W_s^T X) I(W_s^T X; C)}{H(W_s^T X)} \right\rangle \middle| \max_{w_i}\left\langle I(w_i^T X; C) - \sum_{w_s \in W_s} \frac{I(w_i^T X; W_s^T X) I(W_s^T X; C)}{H(W_s^T X)} \right\rangle \right\}$$

   $$W_s \leftarrow \{w_i\}$$

   End for

4) Extract the feature

   $$Y = W_s^T X$$

   $$S \leftarrow \{Y\}$$

5) Output the set S containing the extracted features.

## 4. RESULTS

In this section, we investigate the performance of the proposed method using several UCI data sets (The UCI machine learning repository contains many real-world data sets that have been used by a large variety of investigators) [28] and compare the obtained results with the well-known feature extraction methods: PCA, LDA, SDA and MI-based feature extraction method proposed in known as MRMI-SIG.

A support vector machine (SVM) [29] with a Gaussian radial basis function as a kernel and a KNN classifier [30] has been applied to evaluate the classification performance. SVM classifier is usually picked because it has been proved that it's less sensitive to the curse of dimensionality than other classifiers so the quantity of information that data carry about classes will be in a high correlation with its performance. It's a tenfold cross validation procedure on the training data that is used to determine the cost and the width of the kernel in SVM. Instead KNN classifier with all its simplicity performs so well in the experiments that its often used to compare methods, that's why we have applied KNN with K=1 in this paper. For getting more reliable results, Dividing on absolute maximum of training set normalizes input values of the data for all the classifiers.

For increasing the significance of our result statistics obtained from using data sets with samples of limited number, to obtain the classification rates the average values over 10-fold cross-validation have been applied. To assess the classification accuracy for every 10-fold partition, nine were used as training set and one as test set. First the algorithm of our feature extraction runs on our training sets and then the classifier is trained and tested by them and at the end the average classification results is reported as the error. To evaluate the performance of our method presented in this paper, six data sets have been used. Table I shows brief information of the data sets used in this paper. In the following section we will give a short description of each data set and also the obtained result of examining those using KNN and SVM classifiers and then compare the results with those from the other methods.

Table 1. Description of the data sets used in the comparison.

| Data set | Features | Classes | Samples |
|---|---|---|---|
| Letter | 16 | 26 | 20000 |
| Libras movement | 90 | 15 | 360 |
| Wall-Following | 24 | 4 | 5456 |
| Landsat | 36 | 6 | 6435 |
| Pen based | 16 | 9 | 10992 |
| Image segmentation | 19 | 7 | 2310 |

The first data set used in our paper is Letter data set which its objective is identifying of the 26 capital letters of English alphabet from each other. It is made up of 20000 samples and each sample itself made up of 16 attributes which were then scaled to fit into range of integer values from 0 through 15. Here according to results it is concluded that our method is generally better. In the first five components according to KNN and SVM classifiers our method has the priority of performance but in the last two components both classifiers show that MRMI is better in extracting discriminative information while our method stands in the second place (Table 2).

Table 2. Percentile average classification accuracy on Letter data set.

| KNN | | | | | | | SVM | | | | | | |
|---|---|---|---|---|---|---|---|---|---|---|---|---|---|
| Dim. | Raw | PCA | LDA | SDA | MRMI | MIFX | Dim. | Raw | PCA | LDA | SDA | MRMI | MIFX |
| 1 | 4.4 | 15.2 | 22.1 | 22.2 | 16.3 | **23.2** | 1 | 7.5 | 8.1 | 19.3 | 19.3 | 12.1 | **22.0** |
| 2 | 6.4 | 21.1 | 40.0 | 40.1 | 24.8 | **41.8** | 2 | 8.8 | 19.0 | 43.4 | 43.5 | 26.6 | **46.0** |
| 3 | 10.3 | 33.7 | 51.7 | 51.7 | 40.6 | **59.3** | 3 | 12.8 | 36.4 | 56.0 | 56.0 | 44.8 | **62.1** |
| 4 | 13.6 | 53.8 | 67.0 | 67.0 | 60.0 | **70.9** | 4 | 15.8 | 59.1 | 70.4 | 70.4 | 61.4 | **72.5** |
| 5 | 20.6 | 68.3 | 74.4 | 74.4 | 74.3 | **79.5** | 5 | 23.0 | 72.9 | 76.7 | 76.7 | 77.30 | **80.8** |
| 6 | 30.2 | 77.1 | 81.6 | 81.6 | **84.1** | 83.6 | 6 | 34.3 | 79.8 | 83.7 | 83.7 | **87.0** | 85.4 |
| 7 | 45.9 | 85.9 | 85.8 | 85.8 | **90.7** | 86.3 | 7 | 51.6 | 84.2 | 87.5 | 87.5 | **92.6** | 88.5 |

The second one is Libras movement data set which contains 15 classes and 24 instances where each class refers to a hand movement type in LIBRAS and it represents the coordinates of movements with its 90 features. In this data set KNN and SVM classifiers showed that the results obtained from our method generally is better than the other methods which are as follows: In the first six components KNN and SVM classifier show that our method is better than the other methods but in the seventh component KNN classifier shows that SDA with 85.3% is in the first place when our method is in the second place with 85.0% as SVM still shows that our method is better in results (Table 3).

Table 3. Percentile average classification accuracy on Libras movement data set.

| KNN | | | | | | | SVM | | | | | | |
|---|---|---|---|---|---|---|---|---|---|---|---|---|---|
| Dim. | Raw | PCA | LDA | SDA | MRMI | MIFX | Dim. | Raw | PCA | LDA | SDA | MRMI | MIFX |
| 1 | 24.4 | 23.3 | 31.4 | 26.4 | 20.3 | **33.9** | 1 | 15.3 | 9.4 | 19.2 | 16.4 | 7.5 | **27.5** |
| 2 | 46.1 | 33.1 | 50.0 | 43.6 | 26.7 | **54.4** | 2 | 32.2 | 28.1 | 40.3 | 33.9 | 15.8 | **48.9** |
| 3 | 45.3 | 50.6 | 50.6 | 56.1 | 27.8 | **65.6** | 3 | 33.9 | 45.8 | 48.9 | 44.4 | 21.4 | **63.3** |
| 4 | 46.7 | 65.0 | 55.8 | 67.8 | 28.6 | **75.6** | 4 | 35.3 | 61.9 | 51.4 | 60.0 | 26.1 | **74.7** |
| 5 | 46.1 | 71.7 | 61.9 | 76.4 | 35.3 | **79.4** | 5 | 35.3 | 67.8 | 54.4 | 69.2 | 31.1 | **79.2** |
| 6 | 46.4 | 78.6 | 62.2 | 82.8 | 35.8 | **83.3** | 6 | 35.6 | 71.4 | 55.8 | 73.9 | 34.2 | **79.7** |
| 7 | 45.6 | 81.9 | 63.6 | **85.3** | 36.1 | 85.0 | 7 | 36.7 | 72.5 | 57.5 | 77.8 | 36.4 | **82.8** |

The third is Wall-following robot navigation data set which is made up of 24 features and its objective is testing the hypothesis that this apparently simple navigation task is indeed a non-linearly separable classification task. Therefore linear classifiers unlike non-linear ones cannot be trained to perform navigations around the room without collisions. In this case obtained results from KNN and SVM classifiers show that our feature extraction method is able to obtain better discriminative information than the other five methods in the first seven components which were extracted by them. Other methods could not extract the discriminative information because this data set contains features that are nonlinearly separable (Table 4).

Table 4. Percentile average classification accuracy on Wall-Following data set.

| KNN | | | | | | | SVM | | | | | | |
|---|---|---|---|---|---|---|---|---|---|---|---|---|---|
| Dim. | Raw | PCA | LDA | SDA | MRMI | MIFX | Dim. | Raw | PCA | LDA | SDA | MRMI | MIFX |
| 1 | 50.0 | 39.3 | 49.7 | 45.2 | 40.4 | **61.1** | 1 | 55.7 | 44.5 | 62.3 | 55.8 | 48.3 | **70.4** |
| 2 | 72.2 | 54.1 | 65.9 | 64.0 | 56.5 | **75.5** | 2 | 69.1 | 52.8 | 68.2 | 64.9 | 58.1 | **76.3** |
| 3 | 81.8 | 72.5 | 75.2 | 77.1 | 68.7 | **83.1** | 3 | 79.7 | 67.6 | 73.3 | 75.5 | 66.5 | **80.9** |
| 4 | 85.0 | 80.2 | - | 81.8 | 75.3 | **86.7** | 4 | 83.3 | 76.3 | - | 78.0 | 75.1 | **85.3** |
| 5 | 85.7 | 84.3 | - | 84.7 | 80.0 | **87.4** | 5 | 84.5 | 83.1 | - | 81.8 | 80.2 | **87.2** |
| 6 | 85.2 | 87.2 | - | 85.8 | 82.7 | **88.3** | 6 | 82.8 | 86.7 | - | 83.7 | 82.0 | **88.6** |
| 7 | 84.2 | 87.3 | - | 86.8 | 84.4 | **89.3** | 7 | 80.9 | 86.9 | - | 86.5 | 82.5 | **89.1** |

The fourth one is Landsat satellite data set which is one of the sources of information for a scene consisting of the multi-spectral values of pixels in 3x3 neighborhoods in a satellite image, and the classification associated with the central pixel in each neighborhood. Its objective is the prediction of the classification, given the multi-spectral values. Here obtained results of KNN and SVM classifiers indicate that our feature extraction method can obtain better discriminative information than the other five methods in the first seven components (Table 5).

Table 5. Percentile average classification accuracy on Landsat data set.

| KNN | | | | | | | SVM | | | | | | |
|---|---|---|---|---|---|---|---|---|---|---|---|---|---|
| Dim. | Raw | PCA | LDA | SDA | MRMI | MIFX | Dim. | Raw | PCA | LDA | SDA | MRMI | MIFX |
| 1 | 37.7 | 35.7 | 46.1 | 46.4 | 24.3 | **62.3** | 1 | 39.9 | 38.0 | 50.3 | 51.6 | 20.3 | **69.0** |
| 2 | 67.4 | 76.3 | 70.0 | 75.0 | 29.8 | **77.8** | 2 | 77.4 | 81.7 | 78.1 | 82.2 | 32.6 | **82.6** |
| 3 | 72.6 | 82.2 | 81.6 | 82.3 | 39.8 | **83.6** | 3 | 80.4 | 85.4 | 85.6 | 85.4 | 45.7 | **87.0** |
| 4 | 76.0 | 83.4 | 83.3 | 84.4 | 45.1 | **85.0** | 4 | 82.8 | 86.4 | 86.5 | 86.7 | 53.6 | **87.6** |
| 5 | 78.3 | 84.0 | 82.7 | 83.5 | 56.5 | **85.6** | 5 | 83.8 | 87.7 | 86.4 | 87.1 | 65.7 | **88.3** |
| 6 | 79.7 | 84.6 | - | 84.9 | 67.2 | **85.7** | 6 | 85.3 | 88.0 | - | 88.0 | 74.7 | **88.4** |
| 7 | 80.4 | 84.7 | - | 83.9 | 75.3 | **85.9** | 7 | 85.5 | 88.4 | - | 88.0 | 81.7 | **88.8** |

The fifth data set is Pen-based of recognition of handwritten digits which in it, samples are made up of 250 random digits which are written by 44 writers. Each sample contains 16 features and its objective is to distinguish digits from each other. In this data set in all seven components using

KNN and SVM classifiers, our method indicated the best performance in extracting the features (Table 6).

Table 6. Percentile average classification accuracy on the Pen based data set.

| KNN  |      |      |      |      |      |      | SVM  |      |      |      |      |      |      |
|------|------|------|------|------|------|------|------|------|------|------|------|------|------|
| Dim. | Raw  | PCA  | LDA  | SDA  | MRMI | MIFX | Dim. | Raw  | PCA  | LDA  | SDA  | MRMI | MIFX |
| 1    | 18.3 | 18.6 | 38.6 | 29.0 | 20.0 | **49.8** | 1 | 27.2 | 28.1 | 51.1 | 40.7 | 27.4 | **61.8** |
| 2    | 34.8 | 49.8 | 64.5 | 51.6 | 34.6 | **75.3** | 2 | 47.2 | 61.9 | 73.4 | 61.9 | 46.0 | **82.2** |
| 3    | 47.7 | 77.8 | 77.3 | 68.9 | 48.7 | **86.1** | 3 | 59.4 | 82.7 | 82.3 | 75.7 | 59.2 | **89.0** |
| 4    | 58.2 | 90.6 | 87.4 | 81.8 | 71.8 | **91.9** | 4 | 67.5 | 92.9 | 89.6 | 85.8 | 78.2 | **93.5** |
| 5    | 71.3 | 94.1 | 92.9 | 90.7 | 86.1 | **95.0** | 5 | 77.1 | 95.5 | 94.4 | 92.3 | 89.3 | **96.1** |
| 6    | 80.4 | 95.9 | 96.1 | 94.6 | 93.2 | **96.8** | 6 | 85.1 | 96.8 | 96.8 | 95.6 | 94.6 | **97.6** |
| 7    | 86.1 | 97.4 | 97.2 | 96.7 | 96.2 | **97.6** | 7 | 89.0 | 97.6 | 97.6 | 97.4 | 96.4 | **98.2** |

The sixth data set is Image segmentation data set which is made up of 18 features and describes high-level numeric-valued in 7 classes. Here according to the results SDA is the best one with slightly better results than our method according to the KNN and SVM classifiers. in the first component both classifiers indicate our method is the best one but the results get better for the SDA method in the next two components and we descend to the second place, in the fourth component according to KNN classifier LDA has the priority and we are in the next place again while as SVM classifier SDA has the best performance and we are the second. In the fifth and sixth components according to KNN the best method is SDA, we are the second and LDA stands next but by SVM the better results are for SDA and we take the third place after LDA, in the last component based on KNN results our method with 97.7% is the first method and SDA with 97.4% is the second method while SVM shows that SDA with 96.3% is in the first place and our method with 95.2% stands in the second place (Table 7).

Table 7. Percentile average classification accuracy on Image segmentation data set.

| KNN  |      |      |      |      |      |      | SVM  |      |      |      |      |      |      |
|------|------|------|------|------|------|------|------|------|------|------|------|------|------|
| Dim. | Raw  | PCA  | LDA  | SDA  | MRMI | MIFX | Dim. | Raw  | PCA  | LDA  | SDA  | MRMI | MIFX |
| 1    | 19.7 | 35.7 | 58.2 | 58.3 | 34.8 | **67.8** | 1 | 22.1 | 29.1 | 56.9 | 57.6 | 26.7 | **71.0** |
| 2    | 60.6 | 71.6 | 78.4 | **86.2** | 45.9 | 82.1 | 2 | 60.2 | 68.6 | 78.3 | **86.1** | 45.7 | 82.2 |
| 3    | 60.7 | 76.8 | 93.0 | **95.3** | 59.1 | 94.4 | 3 | 58.7 | 73.7 | 92.7 | **94.8** | 60.0 | 93.5 |
| 4    | 60.8 | 90.6 | **96.7** | 96.4 | 73.7 | 96.0 | 4 | 58.3 | 89.4 | 96.3 | **96.4** | 75.6 | 94.8 |
| 5    | 67.1 | 90.6 | 96.2 | **97.2** | 87.8 | 97.0 | 5 | 63.8 | 8.2 | 95.6 | **96.3** | 88.1 | 95.0 |
| 6    | 67.1 | 94.7 | 95.9 | **97.4** | 93.3 | 97.3 | 6 | 63.8 | 92.9 | 95.1 | **96.4** | 91.4 | 94.8 |
| 7    | 69.7 | 97.1 | -    | 97.4 | 96.0 | **97.7** | 7 | 65.6 | 92.4 | -    | **96.3** | 93.1 | 95.2 |

## 5. CONCLUSIONS

Feature extraction plays an important role in classification systems. In this paper a novel feature extraction method for extracting optimal features based on MI was proposed which is called MIFX. The goal of this method is to create new features from transforming the original features so that maximizes the MI between the transformed features and the class labels and minimizes the redundancy. Since the estimation of MI needs the estimation of multivariate probability density function (pdfs) of the data space and the integration on these pdfs, it always encounters with errors of estimation and also high computational cost. In this paper, by proposing a component-by-component gradient ascent method for feature extraction which is based on one-dimensional MI estimates we try to overcome these problems. At each step, a new feature is created that attempts to maximize the MI between the new feature and the target class and to minimize the redundancy. Only one-dimensional MIs are directly estimated, whereas the higher dimensional MIs are analyzed using the one-dimensional MI estimation. The proposed method was evaluated using six databases of UCI data set. The performed experiments have shown that the MIFX

International Journal of Computer Science & Information Technology (IJCSIT) Vol 4, No 3, June 2012

method is a real competitor to other existing methods. It is most often way better than other methods regarding the results and wherever is not it is close to the best. It is at very high reduction degrees that this method happens to be really effective therefore it is going to be really effective for dimensionality reduction.

## Authors

Ali Shadvar was born in Bandaanzali, Iran in 1981. He received the B.Sc. degree in biomedical engineering from Sahand University of Technology, Tabriz, Iran, in 2005, and the M.Sc. degree in biomedical engineering from Iran University of Science and Technology (IUST), Tehran, in 2010. He is currently an Assistant Professor at the University Islamic Azad University, Langroud, Iran. His research involves data mining, feature extraction and selection, Discriminant analysis, Dimension reduction, pattern recognition, machine learning and rough set.

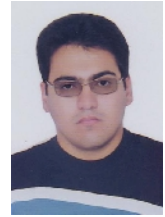